# SO-SLAM: Semantic Object SLAM with Scale Proportional and Symmetrical Texture Constraints

Ziwei Liao, Yutong Hu, Jiadong Zhang, Xianyu Qi, Xiaoyu Zhang, Wei Wang*

*Abstract*— Object SLAM introduces the concept of objects into Simultaneous Localization and Mapping (SLAM) and helps understand indoor scenes for mobile robots and object-level interactive applications. The state-of-art object SLAM systems face challenges such as partial observations, occlusions, unobservable problems, limiting the mapping accuracy and robustness. This paper proposes a novel monocular Semantic Object SLAM (SO-SLAM) system that addresses the introduction of object spatial constraints. We explore three representative spatial constraints, including scale proportional constraint, symmetrical texture constraint and plane supporting constraint. Based on these semantic constraints, we propose two new methods - a more robust object initialization method and an orientation fine optimization method. We have verified the performance of the algorithm on the public datasets and an author-recorded mobile robot dataset and achieved a significant improvement on mapping effects. We will release the code here: https://github.com/XunshanMan/SoSLAM.

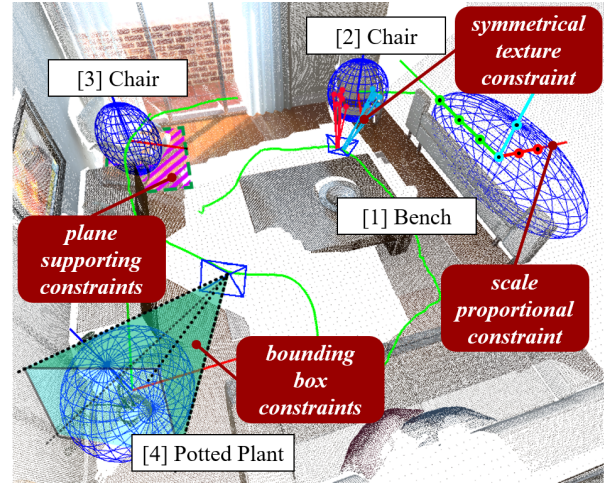

Figure 1. Object SLAM. (*It can build a map with objects including center, orientation and occupied space, to help robots understand object-oriented instructions from human.*)

## I. INTRODUCTION

For decades, robotic researchers have been exploring how to make robots perceive, learn and interact with the environments autonomously in an open world. Imagine a long-term service robot working in an indoor human-robot coexisting scenarios-homes, museums, offices, etc. To respond to human instructions and carry out tasks, it needs the following basic abilities: 1) Robust mapping and localization in disturbed environments. Robots need to robustly detect and locate landmarks under interference from illumination changes, sensor noise, and large viewing angle changes; 2) Modeling, extraction, and reasoning of environmental information in human commands; 3) Environmental changes detection and lifelong map maintaining. The robots need to deal with the random placement of objects such as chairs and teacups, and the addition and removal of furniture.

However, traditional SLAM algorithms use point, line, and plane features to build maps, lacking semantic information [1]. Artificially designed feature descriptors are difficult to adapt to large viewing angle changes and are susceptible to interference from light and sensor noise [2]. The traditional SLAM algorithms are mostly based on static assumptions of the environment. The maps based on points, lines, and planes are difficult to update according to the changes in the environment. Therefore, traditional SLAM algorithms are far from meeting the needs of indoor service robots.

This research is supported by the National Key Research and Development Program of China under grant number 2020YFB1313600.
Authors are with the Robotics Institute, School of Mechanical Engineering and Automation, Beihang University, Beijing, China. Z. Liao completed this work when he was a master candidate at Beihang University and is currently with the Insitute for Aerospace Study (UTIAS) at the University of Toronto, Canada.
* Corresponding author, e-mail: wangweilab@buaa.edu.cn.

We believe that as an important component of the indoor environment, objects have the following potential advantages in the representation of indoor environments:

1) The spatial information of an object can be expressed by more advanced abstract features such as center, orientation, and occupied space. It is a synthesis and high-level version of point, line, and plane features. It is not sensitive to changes in origin observation data and is intuitively more robust for strong disturbances.

2) The spatial relationships between objects and structures (e.g., walls) may be used as auxiliary constraints to improve the robustness and accuracy of object parameters and camera poses estimation. Also, they are conducive to the robot's understanding of the scene from the geometric level to the semantic level, laying a foundation for the robot to execute high-level object-oriented instructions.

Based on the discussion above, we propose a monocular object SLAM system, which builds an object-level map as in Figure 1. We will discuss three representative object spatial constraints: scale proportional constraints, symmetrical texture constraints and plane supporting constraints. We will derive their mathematical representations and constraint models in the SLAM system to participate in both front-end initialization and back-end optimization. Compared with the previous object SLAM systems, we will make the following contributions:

- Propose a monocular object SLAM algorithm that fully couples three spatial structure constraints for indoor environments.

- Propose two new methods based on spatial constraints：a single-frame object initialization method and an object orientation optimization method.

- Testify the effectiveness of the proposed algorithm on two public datasets, and an author-recorded real mobile robot dataset.

## II. RELATED WORK

### A. Object SLAM

Object SLAM, or object-level SLAM, focuses on the construction of object features, including objects' position, orientation, occupied space, and relationships with the spatial structures in the map, as in Figure 1. The early exploration of object SLAM can be traced back to SLAM++ [3]. It establishes an object CAD model database offline, and then uses the depth information of the RGB-D camera to match the object database in actual operation. In 2019, Martin et al. [4] proposed MaskFusion, which uses neural network to detect objects and no database is needed. It detects and tracks dynamic objects in real time. However, limited by their dense object models, the systems mentioned above require heavy computation hardware to achieve real-time and high-rate operations.

In 2019, Yang et al. proposed CubeSLAM [5], which uses cuboids to model objects in the environment. Since 2017, researchers [6][25] have explored the use of quadric models to represent objects in the field of structure from motion. In 2019, Nicholson et al. proposed QuadricSLAM [7], which is the first time to build an object SLAM system with quadrics.

Compared to feature points, cuboids and quadrics can express not only position, but also orientation and occupied space, which are sufficient for robot navigation. Cuboids is a human-defined model, while quadrics have a compact quadratic mathematical representation and complete projective geometry [8]. Recently, the quadric models are getting more and more attention from researchers [9]-[15], and even superquadrics [17] is being explored.

### B. Semantic priors in object SLAM

QuadricSLAM only uses the object detection as the source of observation, making it fragile in the real environment. On the one hand, unlike the ideal looking-around trajectory, the typical straight forward robot moving route is difficult to produce multi-frame observations with large angular variations, raising the problem of unobservability [10]. On the other hand, since the object frame mainly constrains the occupied space, The orientation of the quadric landmark is relatively random and not meaningful.

To make the systems more robust, researchers further explore the meaning of orientation. In [12], gravity and supporting plane are introduced, which defines the "top" side. Ok et al. [10] proposed the texture planes, which essentially defines the "front" side. In addition, deep learning methods are applied to help estimate the ellipsoid [13] as well.

The author has explored the introduction of RGB-D camera for quadrics [14] and tried the object orientation estimation based on symmetry under RGB-D data [15]. This paper will further explore the quadric surface algorithm based on the monocular camera. The monocular algorithm can serve as an effective aid when RGB-D information is missing, ensuring the robustness of the system in real-world scenarios. Also, the monocular camera is more convenient, low-cost, and lightweight, which makes the algorithm available in a wider range of applications, e.g., mobile phones and drones.

In summary, the previous papers are showing that the object SLAM systems based on quadric models are accepted by more and more researchers. However, there is still plenty of room for research to make the systems more robust and accurate in the real world.

## III. MONOCULAR OBJECT SLAM FRAMEWORK

We denote the points in 3D space as set $V = \{v\}$ and the pixels in the image as $U = \{u\}$. The photographic process is denoted as $u = \mathcal{P}(v) = P \cdot v$, where $P$ is the camera projection matrix. An ellipsoid is a point set $V_q = \{v_q \mid \Theta(v_q) = 0\}$, where $\Theta(v) = v^T Q v$. Since $V_q$ is completely determined by $Q$, we can equally call $Q$ an ellipsoid. Representing an object by $Q$ means assuming that all the surface points of the object are on the ellipsoid $Q$.

The front-end input of the system includes the monocular images and odometry data. Object detection algorithm (e.g., YOLO [16]) extracts bounding boxes from RGB images. An ellipsoid $Q$ has 9 degrees of freedom which can be estimated using the SVD method, which needs at least 3 frames of observations with enough view variety [7]. As mentioned in the related work, not only this approach is fragile, but the accuracy of orientation is also lacking.

Human-like orientation perception is necessary to allow service robots to understand and interact with objects. We follow human cognitive habits and consider that the "top" of an artificial object is often the opposite of the supported side of the object, while the "front" is often the direction of symmetry, e.g. cars and chairs. The former defines the direction of the Z-axis and the latter defines the direction of the X-axis, thus the three axes of the object is completely fixed. After that, we can utilize more priories, such as the support relationship and the scale of each direction of the object. Those constraints further allow us to propose an object initialization method that requires only one frame, overcoming the hard-to-meet requirements of SVD method and providing more accurate orientations in the front-end.

In the back end, we model the object SLAM problem as a pose graph, including nodes composed of objects and camera poses, and edges composed of constraints. The object SLAM formulation can be represented as a nonlinear optimization problem:

$$\hat{X}, \hat{Q} = \arg\min_{X,Q} \left( \sum H(F_Z) + \sum H(F_O) + \sum H(F_S) \right) \quad (1)$$

where $X$ is the camera poses and $Q$ is the objects in the map. $F_Z$ is the camera-object observation constraint, $F_o$ is the odometry constraint, and both have been introduced in detail in [7]. This paper emphasizes the newly added $F_s$, composed by plane supporting constraints $f_{sup}$, semantic scale prior $f_{ssc}$ and symmetry constraints $f_{sym}$, which will be introduced in the following parts. $H(\cdot)$ is the robust kernel to make the systems more robust to the outliers, and we use Huber Kernel in the experiments.

## IV. SINGLE-FRAME INITIALIZATION WITH SEMANTIC PRIORS

This paper proposes a method of obtaining 9 degrees of constraints from a single frame observation to initialize a complete ellipsoid. The process is shown in Figure 2. The process will use the following three sources of constraints- object detection constraints, plane supporting constraints and scale proportional constraints. The latter two constraints are spatial structure constraints.

### A. Object detection constraints

As shown in Figure 2, an object $\boldsymbol{O}$ placed on its supporting plane is observed in one frame. The bounding box of the object generated from object detection algorithms in the image is $\boldsymbol{b}$. Generally, the depth and scale of the object is unknown from only one observation. Let the four edges of $\boldsymbol{b}$ be $\boldsymbol{l}_i, i = 1,2,3,4$, then each edge $\boldsymbol{l}_i$ can be back projected to produce a plane $\boldsymbol{\pi}_i$:

$$\boldsymbol{\pi}_i = \boldsymbol{P}^T \boldsymbol{l}_i \quad (2)$$

Each plane will form a tangent constraint with the dual quadrics model $\boldsymbol{Q}^*$ of the object $\boldsymbol{O}$, namely:

$$\boldsymbol{\pi}_i^T \boldsymbol{Q}^* \boldsymbol{\pi}_i = 0, \quad i = 1,2,3,4 \quad (3)$$

So, an object detection constraint $f_{bbox}$ will constitute a constraint of 4 degrees of freedom with the object, which can be expressed as:

$$f_{bbox} = \sum_{i=1} \|\boldsymbol{\pi}_i^T \boldsymbol{Q}^* \boldsymbol{\pi}_i\|_{\Sigma_{det}}, \quad (4)$$

where $\Sigma_{det}$ is the covariance matrix of the bounding box. We use $\Sigma_{det} = 10$ in the experiments.

### B. Plane supporting constraints

In a normal indoor environment, to overcome gravity, objects must form a geometric relationship with spatial structures. E.g., the cup on the table, the lamp under the ceiling, and the paintings on the walls. This paper introduces the most common supporting relationship, when the structural plane is located under the object. The suspension, lean and other relations can be derived in a similar way.

Assuming the supporting plane of the object $\boldsymbol{Q}^*$ is $\boldsymbol{\pi}_s = (\boldsymbol{n}_s, d)$, where $\boldsymbol{n}_s$ is the normal vector of the plane. If the $Z$ axis of the object is upward in the direction of gravity, then its $X$ and $Y$ axes must be orthogonal to the normal vector of the supporting plane, so the following constraints can be obtained:

$$\text{Rot}_x(\boldsymbol{Q}^*) \cdot \boldsymbol{n}_s = 0 \quad (5)$$
$$\text{Rot}_y(\boldsymbol{Q}^*) \cdot \boldsymbol{n}_s = 0 \quad (6)$$

where, $\text{Rot}_x(\boldsymbol{Q}^*)$ is the $X$ axis normal of the ellipsoid $\boldsymbol{Q}^*$. Also, the quadric $\boldsymbol{Q}^*$ should be tangent to the plane $\boldsymbol{\pi}_s$, as:

$$\boldsymbol{\pi}_s^T \boldsymbol{Q}^* \boldsymbol{\pi}_s = 0 \quad (7)$$

So, a supporting plane $\boldsymbol{\pi}_s$ can offer three degrees of constraints to the object $\boldsymbol{Q}^*$ as:

$$f_{sup}(\boldsymbol{Q}^*, \boldsymbol{\pi}_s) = \|Rot_x(\boldsymbol{Q}^*) \cdot \boldsymbol{n}_s\|_{\Sigma_\theta} + \|Rot_y(\boldsymbol{Q}^*) \cdot \boldsymbol{n}_s\|_{\Sigma_\theta} + \|\boldsymbol{\pi}_s^T \boldsymbol{Q}^* \boldsymbol{\pi}_s\|_{\Sigma_\pi}, \quad (8)$$

where, $\Sigma_\theta$ is the rotation covariance, and $\Sigma_\pi$ is the tangent covariance. We use $\Sigma_\theta = \Sigma_\pi = 10$ in the experiments. When the ellipsoid's Z axis is perpendicular to the support plane and its bottom is tangent to the support plane, the constraint error becomes the smallest.

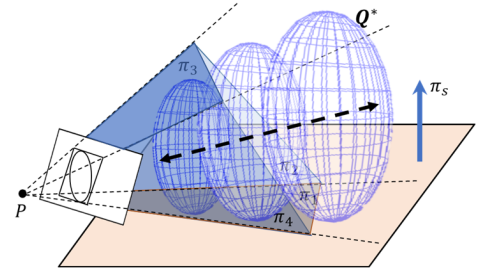

Figure 2. Schematic diagram of the process of recovering the quadric in a single frame. (*The tangent plane of the back projection of the object detection frame and the object supporting surface will jointly constrain the ellipsoid. The depth uncertainty along the observation direction will be further introduced into the scale ratio constraint for estimation.*)

### C. Semantic scale proportional constraint

The scale of indoor artificial objects in the same category has a certain distribution, which is also a geometric reflection of object semantics. There have been some studies discussing how to apply object scale prior constraints to object mapping. Ok et al. [10] assumed that the size of the car is known. However, its flexibility is limited, and it cannot adapt to the scale ambiguity of specific instances with the same label, e.g., a real car and a small toy car.

This paper proposes a new flexible object scale prior- Scale Proportional Constraint (SPC), which constrains object's proportional scale instead of its specific scale. Assuming the scale of an object is $\boldsymbol{s} = [a, b, c]^T$, where $a, b, c$ is the half scale of its $X, Y, Z$ axes. Then we can define its scale ratio $\boldsymbol{r} = [\sigma, \beta]^T$ as follows:

$$\sigma = \frac{a}{c}, \quad (9)$$
$$\beta = \frac{b}{c}, \quad (10)$$

For objects with different semantic labels, a scale ratio table of common objects can be defined, and the ratio can be obtained by querying the table in practical applications. In actual use, the table can be obtained by averaging the scale of common object types.

Given an object $\boldsymbol{Q}_0^*$, its scale ratio $\boldsymbol{r}_0 = r(\boldsymbol{Q}_0^*)$ can be calculated according to the definition. Its corresponding semantic scale ratio $\boldsymbol{r}_s = SemTable(l_0)$ can be obtained by querying the table according to its semantic label $l_0$. Assuming the scale variance is $\Sigma_{ssc}$, the scale ratio constraint of the object $\boldsymbol{Q}_0^*$ with the semantic label $l_0$ is:

$$f_{ssc}(\boldsymbol{Q}_0^*, l_0) = \|\boldsymbol{r}_0 - \boldsymbol{r}_s\|_{\Sigma_{ssc}} \quad (11)$$
$$= \|r(\boldsymbol{Q}_0^*) - SemTable(l_0)\|_{\Sigma_{ssc}},$$

We use $\Sigma_{ssc} = 1$ in the experiments. When the scale value $\boldsymbol{r}_0$ of the object $\boldsymbol{Q}_0^*$ is consistent with its semantic scale prior value $\boldsymbol{r}_s$, the constraint error becomes smallest.

### D. Solving the single frame initialization

Due to the diverse forms of constraints, it is difficult to directly obtain an analytical solution. We construct a nonlinear optimizer, based on the Levenberg-Marquardt algorithm [18], and iteratively solves the optimal value. The objective function is defined as:

$$\hat{Q}^* = \arg\min_{Q^*}(f_{bbox} + f_{sup} + f_{ssc}), \quad (12)$$

including the object detection constraint $f_{bbox}$, the plane supporting constraint $f_{sup}$ and the scale ratio constraint $f_{SSC}$.

## V. ORIENTATION OPTIMIZATION WITH TEXTURE SYMMETRY

### A. Mathematical description of object symmetry

We try to further constrain object orientation by their symmetry property, which is commonly found in man-made objects. The following part of this chapter focuses only on symmetric objects. Geometrically, the *front* of a man-made object is often considered to be the direction of its symmetrical plane. We consider it to correspond to the direction of the X-axis of the object's coordinate system, as in Figure 3(a).

The symmetry of an object is mathematically represented by the fact that for any point $v_0 \in V$ on an object, a point $v_0^S \in V$ can always be found that is symmetric about its plane $\pi_{xz}$. Since the assumption that the object is ellipsoidal, the plane of symmetry can be represented by the elements in matrix $Q$. The symmetry relation between two points about a plane has explicit linear representation [26], denoted as $v_0^S = \mathcal{S}(v_0, Q)$. Considering that objects with multiple symmetry planes such as boxes and balls, we uniformly establish the positive X-axis direction in the direction of the first symmetry surface found when the object is initialized.

For a specific object $Q$, $U_q = \{u_q\}$ is the set of points on the surface of $Q$ in the image plane, then the recovery mapping of the pixel points $u \in U_q$ on the image to the points $v \in V_q$ on the 3D surface of the object is $\mathcal{P}^\dagger(\cdot, Q): U \to V$, such $v$ that $\mathcal{P}^\dagger(u, Q) = v$, then $v$ satisfies:

$$\begin{cases} u = \mathcal{P}(v) = P \cdot v & (a) \\ \Theta(v) = v^T Q v = 0 & (b) \\ v \text{ is visible to camera} & (c) \end{cases} \quad (13)$$

Substituting (a) into (b) yields a quadratic equation about $v$ which has at most one solution subject by (c).

Therefore, for an object point $u_0$ in an image, we can get its symmetry pixel point $u_0^S$:

$$u_0^S = \mathcal{P} \circ \mathcal{S} \circ \mathcal{P}^\dagger(u_0) = \mathcal{P}\left(\mathcal{S}\left[\mathcal{P}^\dagger(u_0, Q), Q\right]\right) := \mathbb{S}(u_0, Q) \quad (14)$$

the process is shown in Figure 3(b), which we write as $\mathbb{S}: U \to U$. Having found the symmetric pixel pairs, we hope to find a descriptor $\beta(\cdot): U \to \mathbb{R}$, to describe the symmetry. Specifically, we want $\beta(\cdot)$ to have the property that:

if $u_1 = \mathbb{S}(u_0, Q_0) = u_0^S$, then $\beta(u_0) = \beta(u_1)$ for all $u_0, u_1$ (15)

When $\beta(\cdot)$ satisfies (15), we say that $\beta(\cdot)$ is **symmetric projection invariant.** After that, we can optimize our ellipsoid $Q$ with cost function $f_{sym}$ when the observation is noisy:

$$f_{sym} = \sum_{u_i} \left(\beta(u_i) - \beta(u_i^S)\right)^2 \quad (16)$$

The next step is to find the descriptor $\beta(\cdot)$

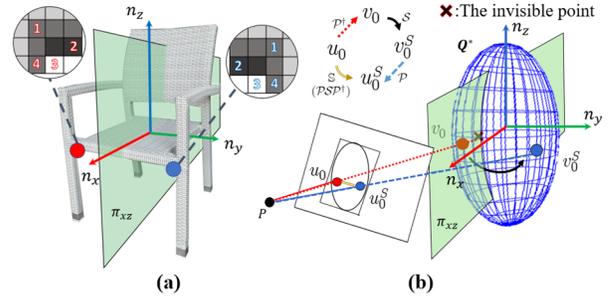

Figure 3. (a) The symmetry relationship between two object points about a plane (b) The process of solving for symmetric pixels in the image.

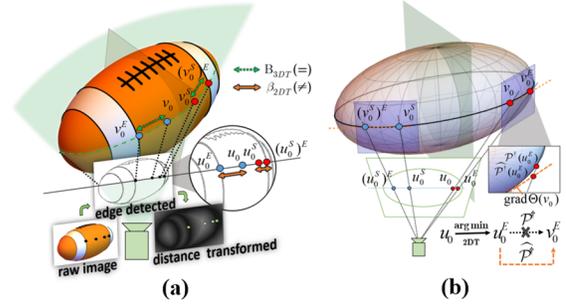

Figure 4. (a) The edge distances of symmetric points are no longer equal after projection distortion. (b) Linearization of the reduction mapping of the edge point.

### B. The construction of symmetry descriptor

Descriptors $\beta(u)$ are needed to reflect some feature of $u$ that is **symmetric projection invariant**. We made different attempts and compared them in section VI.

The most preliminary choice is the grayscale value of the pixel $\beta_{GRAY}(u)$, which, along with its variants, is widely used in the direct-method SLAM, but it is not robust enough in the real situation. Then we tried the BRIEF descriptor $\beta_{BRIEF}(u)$, which can reflect the nearby texture information. To ensure symmetry invariance, the sampling order of texture near $u_0$ and its symmetry point $u_0^S$ should be symmetrical to each other, as in Figure 3(a).

However, in the optimization process (16) using $\beta_{BRIEF}(u)$, with every sampling points $u_i$ fixed, the symmetry points $u_i^S = \mathbb{S}(u_i, Q)$ will change with the optimization iteration of $Q$, and thus $\{u_i^S | i = 1, 2...n\}$ needs to be resampled and recoded in each iteration step, which seriously slows down the algorithm. We then tried to find a more lightweight descriptor to meet the real-time requirements of SLAM which led us to consider the Distance Transform value of pixels.

$$\beta_{2DT}(u_0) = \min_{u \in U_e} \|u - u_0\|, U_e = \{u_e | u_e \text{ is edge pixel}\} \quad (17)$$

The meaning of (17) is *the closest distance from a pixel point to any pixel at the edge of the image*, which partially reflects the object texture. It can be efficiently computed only once for all the pixel points in the object detection frame before the optimization. Then, its value can be **queried** during each iteration.

However, it is doubtful whether the description is symmetric projection invariant. For example, consider the case in Figure 4(a). Note that $v_0^E$ is the nearest edge point of $v_0$, and since symmetric objects have symmetric edge lines, we have $\|v_0 - v_0^E\| = \|v_0^S - (v_0^S)^E\|$. However, affected by projection distortion, the equation no longer holds after projection back to the image, that is, $\|u_0 - u_0^E\| \neq \|u_0^S - (u_0^S)^E\|$, so $\beta_{2DT}(\cdot)$ cannot satisfy (15). Nevertheless, due to the edge symmetry, we found that *the nearest edge distance of point $v_0$*, noted as $B_{3DT}(v_0)$, satisfies

$$\forall v_0, v_1, \text{ if } v_1 = \mathcal{S}(v_0, Q) = v_0^S, \text{ then } B_{3DT}(v_0) = B_{3DT}(v_1) \quad (18)$$

where the definition of $B_{3DT}(\cdot)$ is

$$B_{3DT}(v_0) = \min_{v \in V_e} \|v - v_0\|, V_e = \{\mathcal{P}^\dagger(u_e) \mid u_e \in U_e\} \quad (19)$$

But unlike on the image, the computation of $B_{3DT}(v_0)$ in 3D space needs traversing over every edge point $v \in V_e$ to find the nearest one in each iteration, which makes the computational cost unacceptable again.

Our solution is to combine the advantages of both $\beta_{2DT}(\cdot)$ and $B_{3DT}(\cdot)$, and proposing an **Improved-DT descriptor**. To make it symmetric projection invariant under certain conditions while preserving the lightweight property of *query*, we **assume** that the recovery mapped point $\mathcal{P}^\dagger(u_0^E)$ of the nearest edge pixel $u_0^E$ of pixel $u_0$ in the image **is exactly** the point $v_0^E$ in 3D space, which is the nearest edge point of $v_0 = \mathcal{P}^\dagger(u_0)$. That is

$$v_0^E = \arg\min_{v \in V_e} \|v - v_0\| = \mathcal{P}^\dagger(u_0^E), u_0^E = \arg\min_{u \in U_e} \|u - u_0\| \quad (20)$$

The motivation of (20) is to replace $v_0^E = (\mathcal{P}^\dagger(u_0))^E$ by its approximation $\widehat{v_0^E} = \mathcal{P}^\dagger(u_0^E)$ so that we do not have to iterate over edge points to get the nearest point in 3D space. Further, we can define the descriptor

$$\beta_{3DT}(u_0) = \|\mathcal{P}^\dagger(u_0^E) - v_0\| \approx B_{3DT}(v_0), v_0 = \mathcal{P}^\dagger(u_0) \quad (21)$$

Under assumption (20), $\beta_{3DT}(u_0)$ is **symmetric projection invariant** and can be obtained by simply querying $u_i^E$ and taking $v_i^E = \mathcal{P}^\dagger(u_i^E)$. $u_i^E$ can be obtained by slightly modifying the original distance transform algorithm: Save not only each pixel's shortest distance from the edge but also the corresponding edge pixel coordinate.

### C. Further acceleration of the optimization process

Using the Improved-DT descriptor $\beta_{3DT}(\cdot)$, the cost function is

$$f_{sym}(Q) = \sum_{u_i} \left( \|\mathcal{P}^\dagger(u_i^E) - v_i\| - \|\mathcal{P}^\dagger((u_i^S)^E) - v_i^S\| \right)^2 \quad (22)$$

In each iteration during the optimization, $f_{sym}(Q)$ needs to be recomputed with $Q$ changed and the sampling points $\{u_i\}$ fixed. Although $u_i^S$ still changes with the update, $(u_i^S)^E$ can be obtained directly by query, which significantly speeds up the optimization process. To further accelerate the process, when computing every time-consuming step (13.b) in nonlinear mapping $\mathcal{P}^\dagger(u_0^E)$, since $\Theta(v_0) = v_0^T Q v_0 = 0$ is

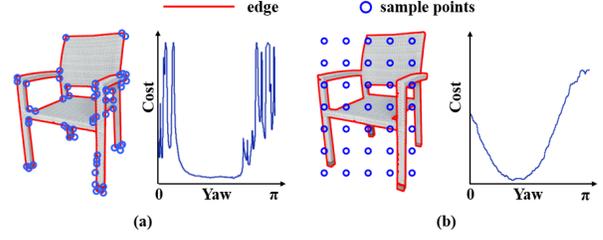

Figure 5. (a) Sampling edge points (left) and its cost function (right) (b) Uniform sampling (left) and its loss function (right)

already satisfied and $u_0^E$ is near $u_0$, consider the linearization of $\Theta(u_0^E) = 0$ at $v_0$:

$$\begin{aligned}
\Theta(v_0^E) &= \Theta\left(v_0 + (v_0^E - v_0)\right) = 0 \\
&= \Theta(v_0) + (v_0^E - v_0)^T \cdot \text{grad}\,\Theta(v_0) + o\|v_0^E - v_0\| \\
&= v_0^T Q v_0 + (v_0^E - v_0)^T \cdot 2Q v_0 + o\|v_0^E - v_0\| \\
&\approx 2(v_0^E)^T Q v_0 = 2\bar{\Theta}_0(v_0^E)
\end{aligned} \quad (23)$$

where $o\|v_0^E - v_0\|$ is the Peano remainder. The geometric meaning of $\bar{\Theta}_0(v_0^E) = (v_0^E)^T \cdot Q v_0 = 0$ is: $v_0^E$ lies on $Q$'s tangent plane $Q v_0$ at $v_0$, as shown in Figure 4(b). Replacing (13.b) by $\bar{\Theta}_0(v_0^E) = 0$ is equivalent to approximating the intersection of a ray with a tangent plane instead of the quadratic surface, whose solution has explicit linear representation [8]. Let the approximate calculation be $\widehat{\mathcal{P}^\dagger}(\cdot)$, then the loss function is

$$f_{sym}(Q) = \sum_{u_i} \left( \|\widehat{\mathcal{P}^\dagger}(u_i^E) - v_i\| - \|\widehat{\mathcal{P}^\dagger}((u_i^S)^E) - v_i^S\| \right)^2 \quad (24)$$

That is, at each iteration, the sampling point mapping $v = \mathcal{P}^\dagger(u)$ is computed accurately and the edge point mapping $v^E \approx \widehat{\mathcal{P}^\dagger}(u^E)$ is computed approximately, which further accelerates of the optimization process.

### D. Strategies of sampling points

We have described in detail the process of constructing the descriptors and how to accelerate the optimization process, leaving only how to obtain the sampling points $\{u_i\}$. Due to remainder $o\|v_0^E - v_0\|$ in (23), the linearization approximation is only valid when $\|v_0^E - v_0\| \approx 0$. Hence, we use two point sampling strategies:

One is to sample the corner points, which can be considered as a stricter edge point, and can guarantee $\|v_0^E - v_0\| \approx 0$. But also because of the nearness, theoretically there will be $\beta_{3DT}(u_i) \equiv 0$, which may cause the gradient vanishing problem near the optimal value, as in Figure 5(a). The other is to uniformly sample points in the bounding boxes. As in Figure 5(b), The gradient problem is significantly improved, but $\|v_0^E - v_0\| \approx 0$ is not quite satisfied. We find

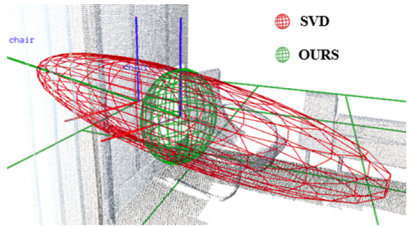

Figure 6. Object initialization result.

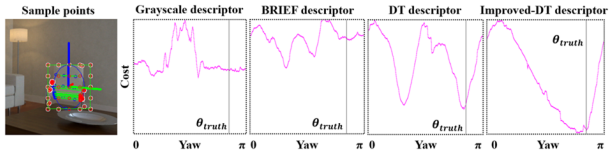

Figure 7. Symmetry cost functions of different descriptors.

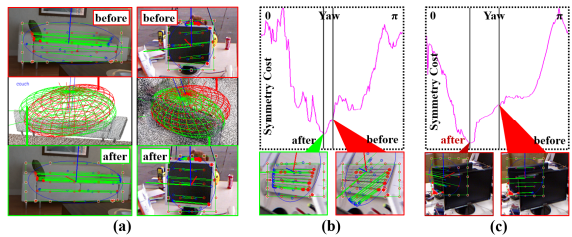

Figure 8. Results before and after symmetry constraints (a) object landmarks (b) cost function value (typical case) (c) cost function value (failure case)

that using corner points together with a few uniform points achieves the best effectiveness in our experiments.

## VI. EXPERIMENTS

### A. Backgrounds

To fully verify the single-frame initialization, texture orientation optimization, and complete system performance proposed in this paper, we conduct experiments both on public datasets, and author-recorded real robot datasets. TUM RGB-D [19] and ICL-NUIM [20] datasets are widely used in SLAM, which cover both room-level and desktop-level environments. To better reflect the effectiveness on mobile robot, we conduct experiments on a turtlebot3 with a Kinect camera operating in a home-like environment, as described in [14]. We use RGB channels only in the experiments. We take every five images to perform object detection with YOLO to get bounding boxes.

We use the indicators IoU and Rot(deg) to fully evaluate the mapping effects. The IoU evaluates the Intersection over Union between their circumscribed cubes of estimated object and ground-truth object. For objects with symmetry, Rot(deg) evaluates the minimum rotation angle required to align the estimated object's three rotation axes with any axis of the ground-truth object to a straight line. For a trajectory, the above metrics are th e average values of all objects' evaluation results. Even though our methods can initialize with only one observation, SVD and QuadricSLAM needs at least three observations. To make the experiments comparable, we consider those objects with at least three observations and filter those partial bounding boxes (those near to the image edges less than 30 pixels), so that all objects can successfully initialize.

TABLE I. SINGLE FRAME OBJECT INITIALIZATION IOU

| Datasets | #Obj | SVD | Cube | Init1-1 | InitP | InitPT |
|---|---|---|---|---|---|---|
| ICL room2 | 4 | 0.072 | 0.33 | 0.363 | **0.484** | 0.478 |
| Fr1_desk | 13 | 0.066 | - | 0.091 | **0.110** | 0.106 |
| Fr2_desk | 12 | 0.130 | - | 0.184 | **0.198** | 0.192 |
| Fr2_dishes | 4 | 0.118 | - | 0.138 | **0.312** | **0.312** |
| Fr3_cabinet | 1 | 0.254 | **0.46** | 0.296 | 0.345 | 0.344 |
| Real-robot | 8 | 0.005 | - | 0.148 | 0.228 | **0.232** |
| Average | 42* | 0.083 | - | 0.163 | **0.218** | 0.215 |

TABLE II. SINGLE FRAME OBJECT INITIALIZATION ORIENTATION

| Datasets | #Obj | SVD | Init1-1 | InitP | InitPT |
|---|---|---|---|---|---|
| ICL room2 | 4 | 32.80 | 17.00 | 15.97 | **13.97** |
| Fr1_desk | 13 | 45.01 | 14.52 | **11.72** | 12.59 |
| Fr2_desk | 12 | 44.67 | 10.93 | 13.44 | **8.80** |
| Fr3_cabinet | 1 | 19.98 | 42.85 | 1.77 | **1.43** |
| Real-robot | 8 | 30.56 | 23.18 | 17.07 | **15.87** |
| Average | 38* | 39.92 | 16.22 | 13.58 | **11.94** |

Since the plane extraction is not our focus, in the experiments, we annotated the support plane in the world coordinate system and then transformed it into the local coordinates of each frame to get the ground-truth planes. In this way, we can know the accuracy limit of our proposed methods. In actual scenarios, support planes can be extracted from point cloud generated from SLAM [2], or directly through a plane SLAM system [21]. For wheeled mobile robots, when considering objects on the ground, the ground plane parameters can be obtained after calibrating the camera's external parameters related to the ground before starting.

### B. Single-frame object initialization

We compare our result with the initialization method SVD of the state-of-art algorithm QuadricSLAM [7] and the initialization method of CubeSLAM [5]. The SVD method requires at least three frames of observations. We put together all observations of the object for SVD initialization in the experiment. Like ours, CubeSLAM introduces the supporting plane to constrain the orientation of the object. We take the experimental results on indoor datasets given in the CubeSLAM paper for comparison. We take the IoU between the estimated object and the ground-truth object as the benchmark, and average over all objects in the trajectory.

Table I and Figure 6 show the results. The SVD method not only requires a larger number of observations, but the accuracy is also lower. Especially in the robot trajectory, the forward motion of the mobile robot is difficult to produce a sufficient perspective difference between observations, resulting in an IoU of only 0.5%. Compared with the SVD method, the initialization of CubeSLAM only needs one observation and obtains better results. CubeSLAM needs extraction of line features to calculate vanish points, which requires the object surface to have obvious straight lines. Ours not only requires one observation, but also has no requirements for the line features of the object. It has a better adaptability to the texture type. Even with the 1:1:1 scale proportional constraint (see Init1-1), Ours achieves an average IoU of 16.3%. With semantic object semantic prior (see InitP), it rises up to 21.8%, which is a significant increase of 13% over SVD. There is also a 15.4% increase over the published data of CubeSLAM on ICL room2. Fr3_cabinet contains a cuboid object only, and CubeSLAM shows the best result.

*C. Orientation estimation based on texture symmetry*

To verify the effectiveness of the Improved-DT descriptor proposed in this paper for representing the symmetry of objects, we analyze the cost compared with Grayscale, BRIEF, and DT descriptors as in Figure 7. The vertical line marks the ground-truth ($\theta_{truth}$) yaw angle of the object in the current frame. The Improved-DT descriptor has an obvious global optimum near the ground-truth value, while others have more than one local optimum. With the change of the object's orientation, its error changes smoother and more significantly. As a result, there are better local gradients to constrain in the optimization process.

Next, we use the Improved-DT descriptor to estimate the orientation of the object as in Figure 8. Table II and IV show the object orientation accuracy after single-frame initialization (see InitPT), and multiple-frames optimization (see OursPT) separately. We only consider the texture constraint of the first observation of each object to avoid constraint conflicts. The orientation result of SVD initialization and QuadricSLAM are given as reference. Although SVD initialization and QuadricSLAM can solve a complete ellipsoid, they do not explicitly constrain the orientation of the object, so the average orientation error is relatively large, reaching 39.9 degrees and 31.7 degrees, respectively. After introducing object supporting constraints and default scale (Init 1-1), real scale proportional prior (InitP) and texture (InitPT), the orientation error was improved and finally achieved 11.9 deg, which is a 64% improvement compared with SVD. With multiple-frame optimization, the orientation (OursPT) was improved to 11.5 deg, and its IoU was increased from 0.215 to 0.286. This accuracy is sufficient for semantic navigation applications involving object orientation, such as commands like "moving to the *front* of the table" and "moving to the *side* of the bench".

We also found some failure cases that could guide future work. On the ICL-NUIM datasets, the orientation accuracy of the large objects such as chair and bench reach about several degree. However, on the fr1_desk and fr2_desk datasets, small objects' orientation such as books and keyboards reach 30-40 degrees. We find that the estimation of the center and scale of the small objects is poor, which makes the three-dimensional symmetry point in the texture constraint not accurate. The occlusion also causes decrease in some clustered environments as in Figure 8(c), which reveals the necessity for multi-frame optimization.

*D. Multiple-frame Optimization*

To compare with the state-of-art algorithms, we reproduced the performance of QuadricSLAM, which is a state-of-art monocular object SLAM system with quadrics. QuadricSLAM uses SVD method to initialize objects and then optimize in the back-end. Data association problem decides which objects in the map an observation belongs to. There has been work [22] concentrating on the problem. Our previous work [14] has also discussed combining quadric model with a nonparametric pose graph to solve the data association. QuadricSLAM paper uses manually annotated data association in the experiments. As data association is not the focus of this paper, we also use manually annotated data association for both QuadricSLAM and ours to testify the best effectiveness in the experiments.

TABLE III. MULTIPLE-FRAME OBJECT ESTIMATION IoU

| Datasets | #Obj | Quadric | OursP | OursPT |
|---|---|---|---|---|
| ICL room2 | 4 | 0.082 | **0.489** | 0.438 |
| Fr1_desk | 13 | 0.071 | **0.135** | 0.131 |
| Fr2_desk | 12 | 0.172 | 0.330 | **0.334** |
| Fr2_dishes | 4 | 0.293 | 0.375 | 0.375 |
| Fr3_cabinet | 1 | 0.255 | 0.316 | 0.317 |
| Real-robot | 8 | 0.034 | 0.308 | **0.343** |
| Average | 42* | 0.120 | 0.285 | **0.286** |

TABLE IV. MULTIPLE-FRAME OBJECT ESTIMATION ORIENTATION

| Datasets | #Obj | Quadric | OursP | OursPT |
|---|---|---|---|---|
| ICL room2 | 4 | 26.96 | 31.60 | 12.23 |
| Fr1_desk | 13 | 36.97 | 14.38 | **13.23** |
| Fr2_desk | 12 | 30.77 | 9.64 | **8.70** |
| Fr3_cabinet | 1 | 18.31 | 1.03 | 0.95 |
| Real-robot | 8 | 28.76 | **13.33** | 13.71 |
| Average | 38* | 31.74 | 14.12 | **11.47** |

TABLE V. RUNNING TIME PERFORMANCE.

| | Init w/o tex | Init w/t tex | Multiple-frames w/o tex | Multiple-frames w/t tex |
|---|---|---|---|---|
| Run Time (ms) | 5.8±1.4 | 116.5±28.4 | 6.9±4.5 | 295.3±230.1 |
| Average Observations | 1 | 1 | 11.8 | 11.8 |

Table III and Table IV show the IoU and orientation accuracy. Both IoU and orientation of QuadricSLAM is improved compared to SVD initialization. With plane supporting constraints and semantic prior, Ours without texture (see OursP) achieved better IoU, orientation of 0.285 and 14.12 deg. After further introducing texture symmetry constraints (see OursPT) on the above basis, the orientation was improved to 11.47 deg, with a slightly improve on IoU. Totally, ours was improved by 138.3% IoU and 63.9% orientation compared with QuadricSLAM. In fr2_dishes and fr3_cabinet, there is smaller gap compared with QuadricSLAM, because in these datasets, the camera trajectory surrounds the object and produces sufficient observations, which is beneficial to QuadricSLAM's optimization.

*E. Computation analysis*

We implemented the algorithm in C++ and used the g2o library for the graph optimization. We present the time-consuming running on a laptop with an Intel Core i5-7200U 2.50GHz CPU, 8GB RAM, on the fr1_desk dataset as in Table V. Note that object detection algorithm is not included. Our algorithm can run in real-time on a common CPU.

*F. Discussion*

The texture orientation constraint is still closely related to the accuracy of the quadric surface itself before optimization. We suppose that decoupling the orientation estimation from other degrees of freedom of the quadrics will have the potential to further improves the orientation estimation effect. We explore several types of symmetry descriptors. Other more complex manually designed descriptors, e.g., FREAK [24], are also potential for estimating symmetry. We leave it as future work to explore their effectiveness with our ellipsoid-based depth estimation. We discuss objects horizontally placed on the support plane in this paper, so it is a valuable future work to explore how to estimate objects pose with 3D rotation.

Trajectories accuracy is an important indicator for SLAM. We did not find significant trajectory accuracy improvements with the introduction of objects. Both QuadricSLAM and our previous work [14] has shown the same conclusion. We think that it is because the odometry data provided by ORB-SLAM2 is already relatively accurate. In low-texture environments such as fr2_dishes and fr3_cabinet, there is more obvious improvement which shows the robustness of object-level features. And we believe that object features' potential is for high-level understandings such as dealing with long-term changes, social navigation, manipulation, etc., instead of localization accuracy.

## VII. CONCLUSION

This paper proposes a monocular object SLAM system, which uses quadrics to model objects and builds an object-level map to represent the environment. This paper introduces three spatial structure constraints- scale proportional constrains, symmetrical texture constrains and supporting plane constrain. Based on these constraints, this paper proposes two new modules- single frame initialization, and orientation fine optimization, which significantly reduces the object SLAM systems' dependences on the number and change of observations. These methods are expected to allow object SLAM to better adapt to the real complex environment. The symmetry constraints on object orientation provides more information for semantic navigation and help the estimation of the scale and center of objects. Considering future work, it will be promising to further explore more types of spatial constraints and semantic priors of objects to help SLAM process.

## REFERENCES

[1] Cadena, Cesar, et al. "Past, present, and future of simultaneous localization and mapping: Toward the robust-perception age." *IEEE Transactions on robotics* 32.6 (2016): 1309-1332.
[2] Mur-Artal, Raul, Jose Maria Martinez Montiel, and Juan D. Tardos. "ORB-SLAM: a versatile and accurate monocular SLAM system." *IEEE transactions on robotics* 31.5 (2015): 1147-1163.
[3] Salas-Moreno, Renato F., et al. "Slam++: Simultaneous localisation and mapping at the level of objects." *Proceedings of the IEEE conference on computer vision and pattern recognition*. 2013.
[4] Runz, Martin, Maud Buffier, and Lourdes Agapito. "Maskfusion: Real-time recognition, tracking and reconstruction of multiple moving objects." *2018 IEEE International Symposium on Mixed and Augmented Reality (ISMAR)*. IEEE, 2018.
[5] Yang, Shichao, and Sebastian Scherer. "Cubeslam: Monocular 3-d object slam." *IEEE Transactions on Robotics* 35.4 (2019): 925-938.
[6] Rubino, Cosimo, Marco Crocco, and Alessio Del Bue. "3d object localisation from multi-view image detections." *IEEE transactions on pattern analysis and machine intelligence* 40.6 (2017): 1281-1294.
[7] Nicholson, Lachlan, Michael Milford, and Niko Sünderhauf. "Quadricslam: Dual quadrics from object detections as landmarks in object-oriented slam." *IEEE Robotics and Automation Letters* 4.1 (2018): 1-8.
[8] Hartley, Richard, and Andrew Zisserman. *Multiple view geometry in computer vision*. Cambridge University Press, 2003.
[9] Gaudillière, Vincent, Gilles Simon, and Marie-Odile Berger. "Camera relocalization with ellipsoidal abstraction of objects." *2019 IEEE International Symposium on Mixed and Augmented Reality (ISMAR)*. IEEE, 2019.
[10] Ok, Kyel, et al. "Robust object-based slam for high-speed autonomous navigation." *2019 International Conference on Robotics and Automation (ICRA)*. IEEE, 2019.
[11] Hosseinzadeh, Mehdi, et al. "Structure aware SLAM using quadrics and planes." *Asian Conference on Computer Vision*. Springer, Cham, 2018.
[12] Jablonsky, Natalie, Michael Milford, and Niko Sünderhauf. "An orientation factor for object-oriented SLAM." *arXiv preprint arXiv:1809.06977* (2018).
[13] Hosseinzadeh, Mehdi, et al. "Real-time monocular object-model aware sparse SLAM." *2019 International Conference on Robotics and Automation (ICRA)*. IEEE, 2019.
[14] Liao, Ziwei, et al. "RGB-D object SLAM using quadrics for indoor environments." *Sensors* 20.18 (2020): 5150.
[15] Liao, Ziwei, et al. "Object-oriented slam using quadrics and symmetry properties for indoor environments." *arXiv preprint arXiv:2004.05303* (2020).
[16] Redmon, Joseph, et al. "You only look once: Unified, real-time object detection." *Proceedings of the IEEE conference on computer vision and pattern recognition*. 2016.
[17] Tschopp, Florian, et al. "Superquadric Object Representation for Optimization-based Semantic SLAM." (2021).
[18] Boyd, Stephen, Stephen P. Boyd, and Lieven Vandenberghe. *Convex optimization*. Cambridge university press, 2004.
[19] Sturm, Jürgen, et al. "A benchmark for the evaluation of RGB-D SLAM systems." *2012 IEEE/RSJ International Conference on Intelligent Robots and Systems*. IEEE, 2012.
[20] Handa, Ankur, et al. "A benchmark for RGB-D visual odometry, 3D reconstruction and SLAM." *2014 IEEE international conference on Robotics and automation (ICRA)*. IEEE, 2014.
[21] Yang, Shichao, et al. "Pop-up slam: Semantic monocular plane slam for low-texture environments." *2016 IEEE/RSJ International Conference on Intelligent Robots and Systems (IROS)*. IEEE, 2016.
[22] Bowman, Sean L., et al. "Probabilistic data association for semantic slam." *2017 IEEE international conference on robotics and automation (ICRA)*. IEEE, 2017.
[23] Thrun, S., & Wegbreit, B. (2005, October). Shape from symmetry. In Tenth IEEE International Conference on Computer Vision (ICCV'05) Volume 1 (Vol. 2, pp.1824-1831). IEEE
[24] Alahi, Alexandre, Raphael Ortiz, and Pierre Vandergheynst. "Freak: Fast retina keypoint." Computer Vision and Pattern Recognition (CVPR), 2012 IEEE Conference on. IEEE, 2012.
[25] Gay, P., Rubino, C., Bansal, V., & Del Bue, A. (2017). Probabilistic structure from motion with objects (psfmo). In Proceedings of the IEEE International Conference on Computer Vision (pp. 3075-3084).
[26] I. Vaisman, *Analytical Geometry*. World Scientific Publishing Company, 1997.